\documentclass[lettersize,journal]{IEEEtran}
\usepackage{amsmath,amsfonts}
\usepackage{algorithmic}
\usepackage{algorithm}
\usepackage{array}
\usepackage[caption=false,font=normalsize,labelfont=sf,textfont=sf]{subfig}
\usepackage{textcomp}
\usepackage{stfloats}
\usepackage{url}
\usepackage{verbatim}
\usepackage{graphicx}
\usepackage{cite}
\usepackage{xcolor}
\usepackage{booktabs}
\usepackage{multirow}
\usepackage{amsmath}
\usepackage{pifont}
\usepackage{hyperref}
\usepackage{bm}
\usepackage{tabularx}
\definecolor{revisioncolor}{HTML}{4169E1} 
  
\allowdisplaybreaks
\begin{document}

\title{SPGrasp: Spatiotemporal Prompt-driven Grasp Synthesis in Dynamic Scenes}

\author{Yunpeng Mei, Hongjie Cao, Yinqiu Xia, Wei Xiao, Zhaohan Feng,  \\ Gang Wang,~\IEEEmembership{Senior Member,~IEEE}, and Jie Chen,~\IEEEmembership{Fellow,~IEEE}
% Fang Deng,~\IEEEmembership{Senior Member,~IEEE}, and Jie Chen,~\IEEEmembership{Fellow,~IEEE}
\thanks{This work was supported in part by the National Natural Science Foundation of China under Grant 62173034, Grant U23B2059, and Grant 62088101.
	
Yunpeng Mei, Hongjie Cao, Yinqiu Xia, Wei Xiao, Zhaohan Feng, and Gang Wang are with the State Key Lab of Autonomous Intelligent Unmanned Systems, Beijing Institute of Technology, Beijing 100081, China (e-mail: meiyunpeng@bit.edu.cn; hongjie.cao@bit.edu.cn; xiayinqiu@bit.edu.cn; xiaowei@bit.edu.cn; zhfeng@bit.edu.cn; gangwang@bit.edu.cn).

Jie Chen is with the Department of Control Science and Engineering, Harbin Institute of
Technology, Harbin 150006, China, and also with the State Key Lab of
Autonomous Intelligent Unmanned Systems, Beijing Institute of Technology,
Beijing 100081, China (e-mail: chenjie@bit.edu.cn).
}}

\maketitle
\begin{abstract}
Real-time interactive grasp synthesis for dynamic objects remains challenging as existing methods fail to achieve low-latency inference while maintaining promptability. To bridge this gap, we propose SPGrasp (spatiotemporal prompt-driven dynamic grasp synthesis), a novel framework extending segment anything model v2 (SAMv2) for video stream grasp estimation. Our core innovation integrates user prompts with spatiotemporal context, enabling real-time interaction with end-to-end latency as low as $59$ ms while ensuring temporal consistency for dynamic objects. In benchmark evaluations, SPGrasp achieves instance-level grasp accuracies of $90.6\%$ on OCID and $93.8\%$ on Jacquard. On the challenging GraspNet-1Billion dataset under continuous tracking, SPGrasp achieves $92.0\%$ accuracy with $73.1$ ms per-frame latency, representing a $58.5\%$ reduction compared to the prior state-of-the-art promptable method RoG-SAM while maintaining competitive accuracy. Real-world experiments involving $13$ moving objects demonstrate a $94.8\%$ success rate in interactive grasping scenarios. These results confirm SPGrasp effectively resolves the latency-interactivity trade-off in dynamic grasp synthesis.
\end{abstract}


\section{Introduction}
\IEEEPARstart{G}{rasp} prediction is a core problem in robotic manipulation and has attracted considerable attention. Most existing approaches perform grasp prediction from single RGB frames, using convolutional or point-based models to infer grasp affordances (e.g., \cite{GRConv,feng2025multi}). 
%GGCNN ,Grasp2024-35
However, relying on isolated frames is inherently limited in dynamic scenarios: it fails to capture temporal cues when objects or the robot itself are in motion  within an otherwise static environment \cite{sam2,Tip,Nips}. Real-world applications underscore the need for robust dynamic grasping, including industrial pipeline sorting \cite{GraspAnything2} and human-robot interaction \cite{universal,manigaussian++}, where robotic perception naturally involves temporal sequences of images.

\begin{figure}[ht] 
	\centerline{\includegraphics[width=1\linewidth]{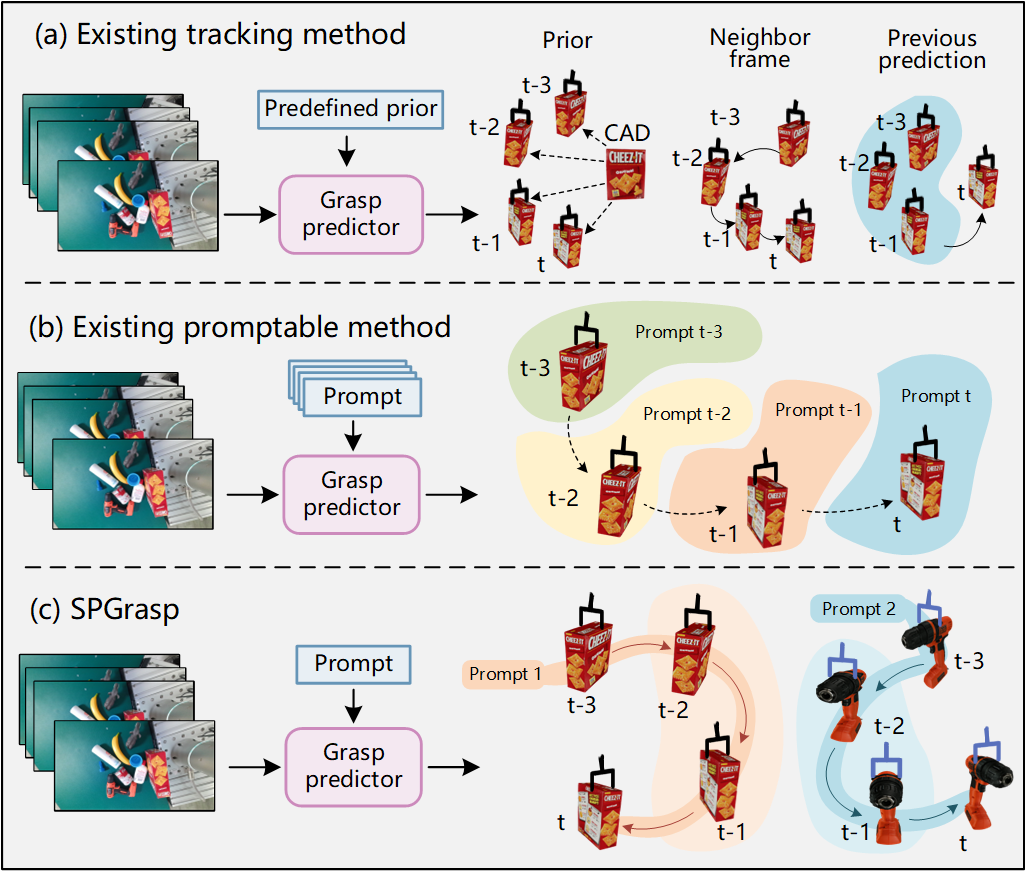}}
	\caption{Architectural comparison: a) Existing grasp tracking methods require object priors (e.g., CAD models) or track predefined instances; b) Promptable grasping approaches need per-frame prompts; c) Proposed SPGrasp tracks prompted objects across frames using only an initial prompt, with light-colored background areas representing the memory features integrating previous prompt, image, and grasp information.}
	\label{fig:related}
\end{figure}

Current methods for dynamic grasp prediction often  decompose the problem by first tracking an object's $6$-DoF pose (using e.g., CAD-model priors \cite{6dtrack}) and then planning graphs based on the tracked pose.  While effective in controlled settings, such model-based tracking approaches require object-specific priors and struggle to generalize to novel (unseen) objects \cite{motion}, which are common in unstructured environments. An alternative paradigm is to predict grasp poses directly from observed video frames, using the history of visual input. For example, some approaches leverage semantic correspondence across adjacent frames for short-term matching \cite{M2}, or predict object trajectories by propagating past grasp predictions \cite{motion}. These methods (see Fig. \ref{fig:related}a) typically assume a predefined set of target objects \cite{Anygrasp}, which leads to a passive response to the environment and prevents user-driven selection of arbitrary objects at runtime. In particular, they cannot flexibly respond to a user’s intention when a specific object is designated by a prompt.

To address these limitations, we introduce SPGrasp, a prompt-driven grasp synthesis framework for dynamic scenes. 
SPGrasp tightly integrates object tracking with interactive prompting by jointly encoding user inputs and spatiotemporal context at each time step. Specifically, SPGrasp accepts multimodal prompts (e.g., language descriptions, bounding boxes, or click points) specified at arbitrary times during a video stream. These prompts are processed via a SAM-inspired architecture \cite{SAM,sam2} to ensure responsive inference. We maintain a spatiotemporal memory bank that accumulates information over time: each frame’s entry stores the frame’s visual features and an object pointer that identifies the tracked target instance. A cross-frame attention mechanism fuses this historical context with the current frame’s features to inform the mask decoder. The model outputs an explicit multi-channel mask with dedicated channels for grasp center, orientation, width, and the object’s semantic mask. These mask channels are decoded into final $4$-DoF grasp poses (center, angle, width) for the targeted object, illustrated in Fig. \ref{fig:framework}. This design enables the system to track and synthesize grasps for the user-specified object in real time, without needing repeated prompting on every frame.

We evaluate SPGrasp on multiple benchmarks. In static, prompt-driven settings (per-frame prompts on single images), SPGrasp achieves $90.6\%$ grasp accuracy on the OCID dataset and $93.8\%$ on Jacquard dataset, which is comparable to state-of-the-art methods. For continuous tracking in dynamic scenes, we train and test on the video-like sequences of GraspNet-1Billion. Here our baseline model achieves $92.0\%$ accuracy overall while running at $73.1$ ms per frame, meeting the accuracy of the best existing promptable methods RoG-SAM \cite{rog-sam} but at a significantly higher speed. In a real-world robotic grasping setup with $13$ moving objects, SPGrasp yields a $94.8\%$ accuracy in an interactive scenario.

The primary contributions of this work are as follows:
\begin{enumerate}
    \item[c1)] \textbf{SPGrasp framework:} We present a novel end-to-end framework for prompt-driven dynamic grasp synthesis from video streams, enabling instance-level grasp pose tracking guided by multimodal user prompts.
    \item[c2)] \textbf{Spatiotemporal context module:} We design a grasp memory module that accumulates historical visual features, explicit grasp predictions, and an object pointer. These components are fused via cross-frame attention, enabling persistent target tracking and consistent grasp prediction in cluttered and occluded scenes, even when only initial prompts are provided.
    \item[c3)] \textbf{Empirical validation:} We conduct extensive experiments on standard benchmarks and in real-world scenarios. SPGrasp achieves comparable accuracy to prior promptable approaches, with $90.6\%$ on OCID, $93.8\%$ on Jacquard, and $92.0\%$ on GraspNet-1Billion, while operating in real time at $73.1$ ms per frame. This represents a $2.4\times$ speed advantage over RoG-SAM. Real-world validation demonstrates a $94.8\%$ success rate for dynamic prompt-guided grasp prediction tasks.
\end{enumerate}

\section{Related Work}
\subsection{Dynamic Grasping}

\begin{figure*}[ht] 
	\centerline{\includegraphics[width=1\linewidth]{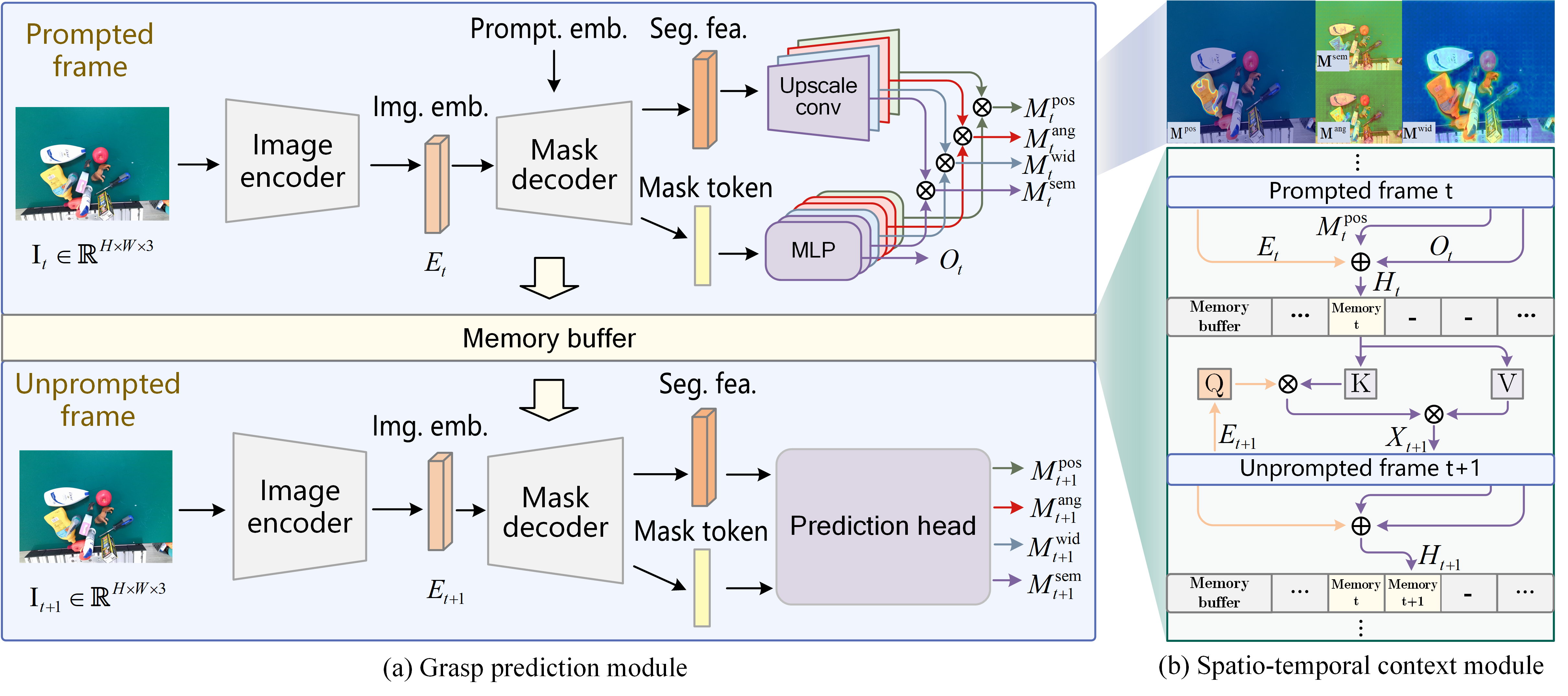}}
    % \vspace{-6pt}
	\caption{SPGrasp framework overview. a) Multi-head grasp prediction pipeline depicting forward data flow, with inter-frame memory buffers integrating historical prompt, image, and grasp features. b) Spatiotemporal context module architecture detailing memory buffer operations and cross-frame attention mechanisms for feature propagation.}
    % \vspace{-2pt}
	\label{fig:framework}
\end{figure*}

Static-scene grasp prediction is well-studied (e.g., \cite{GGCNN,GRConv,Graspnet}), but grasping in dynamic settings—where the robot and objects may be in motion—poses additional challenges, especially under stringent latency constraints \cite{motion}. Early approaches to dynamic grasping decoupled motion tracking from grasping: first estimating object motion or velocity \cite{M2.1,VFAS24} then planning an adaptive grasp trajectory. However, as illustrated in Fig. \ref{fig:framework}(a), these methods typically require object-specific priors (such as CAD models) to track or predict motion, limiting their applicability to novel objects.

More recent works predict grasps directly from video sequences without explicit pose tracking. Some enforce temporal consistency by correlating features or grasps across adjacent frames 
\cite{M2,Anygrasp}. While this improves stability, simple two-frame correspondences can accumulate drift over long sequences. Others use sliding windows of past grasp predictions to inform current decisions \cite{motion}. Although these methods better capture longer-term trends, their reliance on high-level grasp pose history may overlook fine-grained image features (and ignore user input), which can lead to errors in complex scenes.

In contrast, SPGrasp requires no object priors. It embeds user prompts directly into a spatiotemporal memory and leverages multiple context cues: i) low-level image features from past frames, ii) explicit past grasp masks, and, iii) an object pointer for identity. This rich representation preserves the target’s identity over time, mitigating confusion in clutter or under occlusion, and maintains coherent predictions even with sparse prompting.

\subsection{VLM-based Promptable Grasping}
Selecting the intended object is crucial for reliable grasping in clutter. Early instance-level grasping pipelines performed object detection (e.g., YOLO \cite{yolov3,liu2023edgeyolo}) to find candidate targets, then computed grasps for high-confidence detections. These multi-stage methods can succeed in isolated cases, but lack tight integration of user guidance and struggle with open-vocabulary object descriptions.

Vision-language and vision-prompting models have enabled interactive, prompt-driven grasping. For instance, some works use CLIP \cite{CLIP} to map language instructions to visual tokens for grasp synthesis \cite{LanguageGrasp}. 
Others introduced direct visual prompting: for example, 
\cite{prompt2} combines zero-shot segmentation with classification to localize objects from images, but its reliance on exemplar query images and coarse masks limits generality. 

A recent trend is end-to-end models built on large foundation models. Works like RoG-SAM \cite{rog-sam} integrate SAM and Grounding DINO \cite{DINO} to jointly perform instance selection and grasp prediction. These models accept multimodal prompts (points, boxes, text) and eliminate the need for separate detectors. However, all current promptable methods share a limitation: they require a fresh prompt on every frame to guide prediction. This per-frame prompting is acceptable for static or single-image tasks, but impractical in dynamic scenes where continuous user interaction is not feasible.

SPGrasp overcomes this challenge by encoding prompt information in its temporal memory, allowing prompts given at arbitrary frames to persistently influence future frames. In practice, a single prompt can initiate tracking: subsequent frames are handled without user input unless the user issues a new prompt. This capability distinguishes SPGrasp from prior prompt-driven models and enables continuous, interactive grasping in dynamic environments.

\section{Methodology}
\label{sec:methodology}

We now describe the SPGrasp framework for prompt-driven dynamic grasp synthesis. We first formalize the problem (Section~\ref{sec:problem_statement}), then present the architecture (Section~\ref{sec:sp_grasp_architecture}) and training loss (Section~\ref{sec:loss_function}).

\subsection{Problem Statement}
\label{sec:problem_statement}
Grasping specific target objects in video streams poses significant challenges for robotics, particularly in dynamic and unstructured environments. This necessitates a system capable not only of identifying viable grasp affordances but also of consistently tracking and predicting grasp poses for a designated target instance as it evolves temporally. The complexity is further amplified by the requirement for interactive adaptation to user guidance, where a user can specify or update the target object instance at any frame.

Formally, given a video sequence \(\{I_1, I_2, \dots, I_K\}\) with RGB frames \(I_t \in \mathbb{R}^{H \times W \times 3}\) and corresponding user prompts \(\{P_1, P_2, \dots, P_K\}\) (\(P_t\) may be null when leveraging prior context), SPD-Grasp leverages the current frame \(I_t\), current prompt \(P_t\), and an aggregated historical context \(\bar{H}_{t-1}\) from multiple past frames to generate temporally consistent grasp poses per frame. The model first produces an intermediate multi-channel grasp mask \(M_t \in \mathbb{R}^{H \times W \times C}\) providing dense pixel-level predictions of target instance segmentation and grasp attributes (affordance, orientation, width). This mask is decoded into final grasp poses \(\mathcal{G}_t = \{G_{t,j}\}_{j=1}^{N_t}\), where each grasp configuration \(G_{t,j} = (x_{t,j}, y_{t,j}, \theta_{t,j}, w_{t,j})\) defines center coordinates, gripper orientation, and aperture width. This parameterization forms a practical $4$-DoF representation suitable for video-based grasping, as established in prior work~\cite{GraspRPN}. Here, \(N_t\) is the number of grasp poses predicted for the target in frame \(I_t\).

Central to this framework is the formulation and utilization of aggregated historical context. This mechanism is vital for generating temporally coherent grasp predictions and enabling accurate instance tracking from sparse prompts. Section~\ref{sec:sp_grasp_architecture} details SPD-Grasp's architecture for generating \(M_t\) and \(\mathcal{G}_t\) through explicit modeling of these interactions.

\subsection{Architecture of SPGrasp}
\label{sec:sp_grasp_architecture}
SPD-Grasp is engineered to address the aforementioned challenges through a recursive estimation process that leverages multi-frame history. As depicted in Fig.~\ref{fig:framework}, our architecture integrates two core components: a grasp prediction module and a spatiotemporal context module. The prediction process for the intermediate mask \(M_t\) at timestep \(t\) is governed by our core model \(\mathcal{F}\), abstracted as:
\begin{align}
    M_t &= \mathcal{F}(X_t, P_t; \bm{\phi}) \label{eq:mask_prediction_simple} \\
    X_t &= \text{CrossAttn}(\bar{H}_{t-1}, E_t; \bm{\varphi}) \label{eq:fused_features_simple} \\
    H_t &= \text{Concat}(E_t, M_t^{\text{pos}}, O_t) \label{eq:history_update_simple}
\end{align}
where \(M_t\) denotes the multi-channel instance-specific mask generated by model \(\mathcal{F}\) with parameters \(\bm{\phi}\); \(X_t\) represents the spatio-temporal feature fusion of aggregated history \(\bar{H}_{t-1}\) and current visual embedding \(E_t\) via cross-attention mechanism \(\text{CrossAttn}(\cdot; \bm{\varphi})\); \(\bar{H}_{t-1}\) is the aggregated historical context constructed by concatenating the \(N_{\text{hist}}\) most recent state vectors \(\{H_{t-1}, \dots, H_{t-N_{\text{hist}}}\}\); \(P_t\) is the user-provided prompt at timestep \(t\), where non-null prompts initiate new tracking sequences causing model reset to current frame embedding (\(X_t=E_t\));  \(H_t\) is the current state vector formed by concatenating visual embedding \(E_t\), grasp position mask \(M_t^{\text{pos}}\), and object pointer \(O_t\), stored for future aggregation.

\subsubsection{Grasp Prediction Module}
\label{sec:grasp_prediction_module}

The grasp prediction module generates the multi-channel, instance-specific mask \(M_t\). Its architecture, illustrated in Fig.~\ref{fig:framework}(a), draws inspiration from SAM \cite{SAM,sam2} and comprises three core components: an image encoder, a prompt encoder, and a mask decoder.

The image encoder transforms input frame \(I_t\) into visual embedding \(E_t\), while the prompt encoder processes user-provided prompt \(P_t\) into an internal feature representation. The mask decoder synthesizes multi-channel mask \(M_t\) by conditioning on feature representation \(X_t\) and the processed prompt, where \(X_t\) corresponds to pure visual embedding \(E_t\) for prompted frames or fused spatio-temporal features for unprompted frames.

The module operates in two distinct modes based on prompt availability: For prompted frames (\(P_t \neq \text{null}\)), new prompts initiate tracking sequences with predictions conditioned solely on current visual information (\(X_t=E_t\)), intentionally disregarding history to enable re-targeting. For unprompted frames (\(P_t = \text{null}\)), tracking mode leverages fused spatio-temporal features \(X_t\) incorporating aggregated historical context \(\bar{H}_{t-1}\) to integrate spatiotemporal grasp information.

The output \(M_t\) is a \(C=5\) channel tensor: Channel 1 represents grasp position mask \(M_t^{\text{pos}}\). Channels 2-4 encode grasp geometry (\(\sin(2\theta)\), \(\cos(2\theta)\), and grasp width mask) decoded into final $4$-DoF grasp poses \(\mathcal{G}_t\). Channel 5 provides semantic segmentation mask \(M_t^{\text{sem}}\) of the target object, essential for deriving object pointer \(O_t\).

\subsubsection{Spatiotemporal Context Module}
\label{sec:spatio_temporal_context}
The spatio-temporal context module enables SPD-Grasp to maintain consistent instance tracking across unprompted video segments. As depicted in Fig.~\ref{fig:framework}(b), this module manages historical information and integrates it with current frame data through three key mechanisms: a FIFO memory buffer, an object pointer, and a cross-frame attention mechanism.

\textbf{FIFO memory buffer.} 
A first-in-first-out buffer of size \(N_{\text{hist}}\) forms the temporal modeling core. Upon receiving new prompts, the buffer resets to initialize new tracking sequences; for subsequent unprompted frames, it provides recent state vectors \(\{H_{t-1}, \dots, H_{t-N_{\text{hist}}}\}\) to construct aggregated historical context \(\bar{H}_{t-1}\).

\textbf{Object pointer.}
This compact representation derived from mask token of the multi-channel mask \(M_t\) and maintains persistent object identity across frames. Stored in state vector \(H_t\) and the FIFO buffer, it ensures tracking correspondence: For prompted frames, user prompt \(P_t\) directly guides prediction of \(M_t\) and derivation of \(O_t\), registering the target instance. For unprompted frames, the model infers object grasp from aggregated history \(\bar{H}_{t-1}\) to predict \(M_t\) and propagate \(O_t\).

\textbf{Cross-frame attention.}
This mechanism integrates historical context with current visual data during unprompted tracking. Taking aggregated history \(\bar{H}_{t-1}\) and current embedding \(E_t\) as inputs, it computes fused feature representation \(X_t\) by using \(E_t\) as query to attend to keys/values from \(\bar{H}_{t-1}\). The resulting representation captures relevant historical cues and feeds into the mask decoder for predictions informed by current and historical data.

\subsection{Loss Function}
\label{sec:loss_function}
SPGrasp is trained using a multi-component loss function \(\mathcal{L}_{\text{total}}\) applied to the sequence of predicted multi-channel masks for each video clip. For any frame in the sequence, the predicted mask \(M \in \mathbb{R}^{H \times W \times C}\) (with \(C=5\)) contains five channels representing distinct grasp attributes: channel $1$ (grasp location), channels $2$ and $3$ (cosine and sine of \(2\theta\), respectively, for grasp angle), channel $4$ (grasp width), and channel $5$ (semantic object mask). We denote the prediction for channel \(c\) of a given frame as \(M_c \in \mathbb{R}^{H \times W}\), with corresponding ground truth \(T_c \in \mathbb{R}^{H \times W}\).
The overall loss comprises four weighted components: position loss (\(\mathcal{L}_{\text{pos}}\)), angle loss (\(\mathcal{L}_{\text{ang}}\)), width loss (\(\mathcal{L}_{\text{wid}}\)), and semantic mask loss (\(\mathcal{L}_{\text{semantic}}\)), each scaled by importance factor \(w_c\). The weights are assigned as: \(w_1=5.0\) (position), \(w_2=w_3=5.0\) (angle components), \(w_4=1.0\) (width), and \(w_5=1.0\) (semantic mask).
The loss \(\ell_c\) for each channel \(c\) of a given frame is computed as follows.

\paragraph{Binary attribute loss}
For channels \(c \in \{1, 4, 5\}\) (position, width, semantic mask), we employ a binary cross-entropy loss with class-balancing. Positive pixels are up-weighted by factor \(\alpha_c\), dynamically calculated as the negative-to-positive sample ratio in target \(T_c\) and clamped by channel-specific maximum \(\beta_c\). The clamping values are \(\beta_1=20\), \(\beta_4=10\), and \(\beta_5=5\). The loss is defined as
\begin{equation}
\begin{split}
	\ell_c &= -\frac{1}{HW} \sum_{i,j} \Big[ 
		\alpha_c T_{c,ij} \log(\sigma(M_{c,ij})) \\
		&\quad + (1 - T_{c,ij}) \log(1 - \sigma(M_{c,ij})) 
	\Big]
\end{split}
	\label{eq:bce_loss_simplified}
\end{equation}
where \(\sigma(\cdot)\) is the sigmoid function, summing over all pixels \((i,j)\).

\paragraph{Grasp angle loss}
For angle channels \(c \in \{2, 3\}\), where predictions \(M_c\) are constrained to \([-1, 1]\), loss \(\ell_c\) computes the mean-squared error (MSE) between prediction \(M_c\) and target \(T_c\):
\begin{equation}
	\ell_c = \frac{1}{HW} \sum_{i,j} (M_{c,ij} - T_{c,ij})^2
	\label{eq:angle_loss_mse_simplified}
\end{equation}
summed over all pixels \((i,j)\).

\paragraph{Aggregation of losses}
Per-frame component losses aggregate weighted channel-specific losses:
\begin{subequations}
  \begin{align}
	\mathcal{L}_{\text{pos}} &= w_1 \ell_1 \label{eq:L_pos_comp_simplified} \\
	\mathcal{L}_{\text{ang}} &= w_2 \ell_2 + w_3 \ell_3 \label{eq:L_ang_comp_simplified} \\
	\mathcal{L}_{\text{wid}} &= w_4 \ell_4 \label{eq:L_wid_comp_simplified} \\
	\mathcal{L}_{\text{semantic}} &= w_5 \ell_5 \label{eq:L_sem_comp_simplified}.
\end{align}  
\end{subequations}

Each training sample contains a video clip of fixed length \(N_\text{clip}\). The final objective \(\mathcal{L}_{\text{total}}\) averages the frame-level losses across the clip:
\begin{equation}
    \mathcal{L}_{\text{total}} = \frac{1}{N_{\text{clip}}} \sum_{t=1}^{N_{\text{clip}}} \left( \mathcal{L}_{\text{pos}, t} + \mathcal{L}_{\text{ang}, t} + \mathcal{L}_{\text{wid}, t} + \mathcal{L}_{\text{semantic}, t} \right).
    \label{eq:l_total_final_simplified}
\end{equation}

\section{Experiment}
\subsection{Grasp Datasets}
To evaluate the performance of SPGrasp, we selected three public datasets, each chosen to assess a distinct aspect of our model's capabilities. We utilize the Jacquard dataset to test grasp pose generalization across a wide variety of objects, the OCID dataset to evaluate prompt-driven instance selection in cluttered scenes, and the GraspNet-1Billion dataset to benchmark promptable dynamic grasp prediction.

\textbf{OCID:}
The OCID \cite{OCID} focuses on object recognition and grasping in cluttered indoor environments. It provides instance-level labels for objects within complex arrangements, making it particularly well-suited for evaluating the ability of SPGrasp to select and generate grasps for a specific, user-prompted object amidst distractors. The dataset includes over $11,000$ images and $75,000$ annotated grasp poses.

\textbf{Jacquard:}
The Jacquard dataset \cite{Jacquard} is a large-scale collection known for its vast object diversity. Featuring over $11,600$ unique objects across approximately $54,000$ images and $1.1$ million grasp annotations, it serves as an ideal benchmark for assessing the generalization performance of our model. We use this dataset to test the model's ability to predict valid grasp poses for previously unseen objects.

\textbf{GraspNet-1Billion:}
GraspNet-1Billion \cite{Graspnet} provides a comprehensive benchmark for object grasping in realistic cluttered environments. It includes $190$ scenes with $97,280$ RGB-D images, spanning $88$ distinct objects and over $1.1$ billion annotated grasp poses. Critically, data collection involved camera movement through predefined waypoints, providing sequential scene views. This structure enables treatment as video streams, making it ideal for evaluating SPGrasp's dynamic grasp synthesis and temporal tracking. For experiments, we construct video clips by sampling \(N_\text{clip}\) consecutive frames from these sequences.

Across all datasets, SPGrasp was initialized with pre-trained weights from the SAM2-Base model \cite{sam2}. Fine-tuning and evaluation were performed using four NVIDIA GeForce RTX 4090 GPUs, employing the Adam optimizer with an initial learning rate of \(5.0 \times 10^{-4}\). For OCID and Jacquard, training used a batch size of $4$ over $80$ epochs. For GraspNet-1Billion (with video clip inputs), a batch size of $2$ and $200$ epochs ensured comprehensive frame coverage.

The performance of SPgrasp is quantified using Jaccard index and angle difference metric. A prediction is considered correct if it satisfies two conditions. First, the deviation between the predicted and ground-truth grasp orientations is less than \(30^{\circ}\). Second, the intersection over union (IoU) between the predicted grasp rectangle \(G_\text{pred}\) and the corresponding ground-truth rectangle \(G_\text{truth}\) must be greater than $0.25$. The IoU is formally defined as:
\begin{equation}
	\text{IoU}(G_\text{pred}, G_\text{truth}) = \frac{|G_{\text{pred}} \cap G_{\text{truth}}|}{|G_{\text{pred}} \cup G_{\text{truth}}|}.
	\label{eq:iou}
\end{equation}

\begin{table}[h]
	\centering
	\caption{Quantitative comparison between SPGrasp and existing methods on OCID and Jacquard benchmarks}
	\newcommand{\thicktoprule}{\toprule[1pt]}
	\begin{tabular}{c c c c c c}
		\thicktoprule
		{Method} &{Input} &{IL} & {Prompt} &{OCID (\%)} &{Jacquard (\%)} \tabularnewline
		\midrule
        GR-ConvNet \cite{GRConv} 	& RGBD 	&\ding{55}	&\ding{55} 	&$-$  &$\mathbf{94.6}$\tabularnewline
        Grasp-Det \cite{OCID} 		& RGB	&\checkmark	& \ding{55}	&$89.0$  &$92.9$ \tabularnewline
        SSG	\cite{Ins} 			& RGBD	&\checkmark	&\ding{55} & $\mathbf{92.0}$  & $91.8$ \tabularnewline
        RoG-SAM \cite{rog-sam}		& RGB &\checkmark	&\checkmark   & $90.1$  & $\mathbf{94.6}$ \tabularnewline
        SPGrasp          & RGB & \checkmark  &   \checkmark  & $\underline{90.6}$ & $\underline{93.8}$ \tabularnewline
		\bottomrule
	\end{tabular}
	\label{table:jacquard}
\end{table}

 \begin{figure*}[ht] 
	\centerline{\includegraphics[width=1\linewidth]{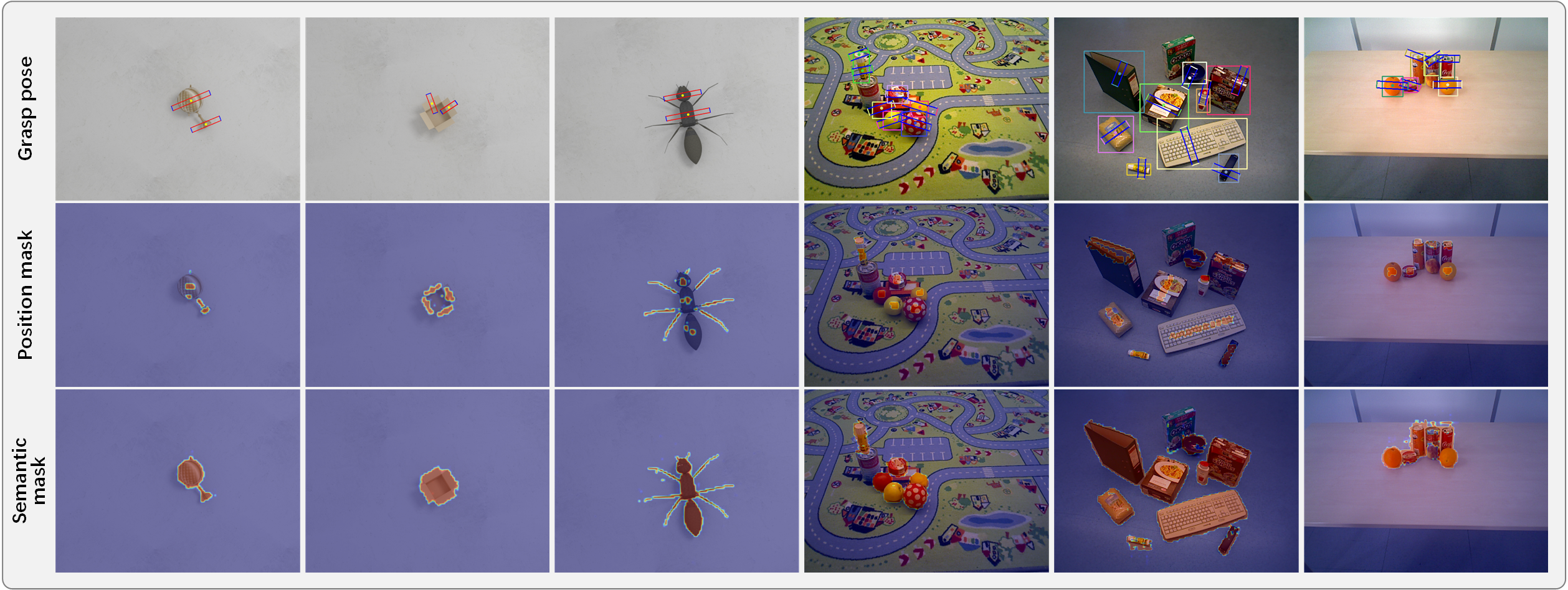}}
	\caption{Qualitative grasp predictions by SPGrasp on Jacquard and OCID Datasets. Left: Jacquard examples (three columns). Right: OCID examples (three columns). For each dataset,  top row: Input image with predicted grasp poses (multi-box prompts visualized for OCID); middle row: Grasp position mask heatmap; bottom row: Object semantic segmentation mask.}
	\label{fig:jacquard_OCID}
\end{figure*}
\subsection{Experimental Results}
To evaluate SPGrasp's generalization capability across diverse scenes and objects, we conducted prompt-based grasp prediction experiments on OCID and Jacquard datasets. Comparative results reported in Table \ref{table:jacquard} focus on diverse static grasp prediction, where each frame receives a user prompt.
It is clear that SPGrasp achieves grasp accuracies of $90.6\%$ on OCID and $93.8\%$ on Jacquard. The table categorizes compared methods by input modality, instance-level grasping capability, and promptability. These results demonstrate that in scenarios without spatiotemporal context, SPGrasp performs comparably to state-of-the-art prompt-driven methods like RoG-SAM. Qualitative predictions from both methods are shown in Fig. \ref{fig:jacquard_OCID}, displaying results for Jacquard (first three columns) and OCID (last three columns). Each dataset presents:
\begin{enumerate}
    \item Input image with predicted grasp poses (rectangles), including multi-box prompts on OCID;
    \item Heatmap of predicted grasp position mask (channel $1$ of \(M_t\));
    \item Predicted semantic segmentation mask of target object (channel $5$ of \(M_t\)).
\end{enumerate}
As illustrated, SPGrasp effectively distinguishes between a target's broader semantic region and its specific graspable locations (inferred from physical affordances), generating feasible grasp poses where the graspable area constitutes a subset of the semantic mask.

To evaluate SPGrasp's performance in dynamic scenes, we conducted experiments on the GraspNet-1Billion dataset, comprising $190$ video sequences ($256$ frames each). Our evaluation examined the impact of key hyperparameters, specifically history length \(N_\text{hist}\) and clip length \(N_\text{clip}\), while comparing against relevant grasp prediction methods. Our baseline SPGrasp configuration used \(N_\text{hist} = 8\) and \(N_\text{clip} = 8\) with a SAM2-Base backbone, balancing performance and training efficiency.

\begin{table*}[htbp] 
\centering
\caption{Benchmark Results Comparing SPGrasp with Existing Grasp Detectors on GraspNet-1Billion}
\label{table:graspnet1}
\begin{tabular}{ccccccccc}
\toprule
Method & Input & IL & Prompt & Interval (fr.) & Seen (\%) & Similar (\%) & Novel (\%) & Overall (\%) \\ 
\midrule
Multi-grasp \cite{multigrasp} & RGBD & \ding{55} & \ding{55} & - & $82.7$ & $77.8$ & $72.7$ & $77.7$  \\
GG-CNN2 \cite{Morrison} & D & \ding{55} & \ding{55} & - & $83.0$ & $79.4$ & $76.3$ & $79.6$  \\
MSGNet \cite{MSGNet} & RGBD & \checkmark & \ding{55} & - & - & - & - & $89.7$  \\
GR-ConvNet2 \cite{GRConv2} & RGBD & \ding{55} & \ding{55} & - & \textbf{97.9} & \textbf{96.0} & $90.5$ & \textbf{94.8}  \\
RoG-SAM \cite{rog-sam} & RGB & \checkmark & \checkmark & $1$ & \underline{93.6} & $90.9$ & $89.1$ & $91.2$  \\
\midrule 
SPGrasp & RGB & \checkmark & \checkmark & $1$ & $92.2$ & $91.5$ & \textbf{92.0} & $91.9$ \\
SPGrasp & RGB & \checkmark & \checkmark & 8 & $92.5$ & \underline{91.7} & \underline{91.9} & \underline{92.0} \\
SPGrasp & RGB & \checkmark & \checkmark & $256$ & $88.3$ & $86.4$ & $86.8$ & $87.2$ \\
\bottomrule
\end{tabular}
\end{table*}

\begin{figure*}[ht] 
	\centerline{\includegraphics[width=1\linewidth]{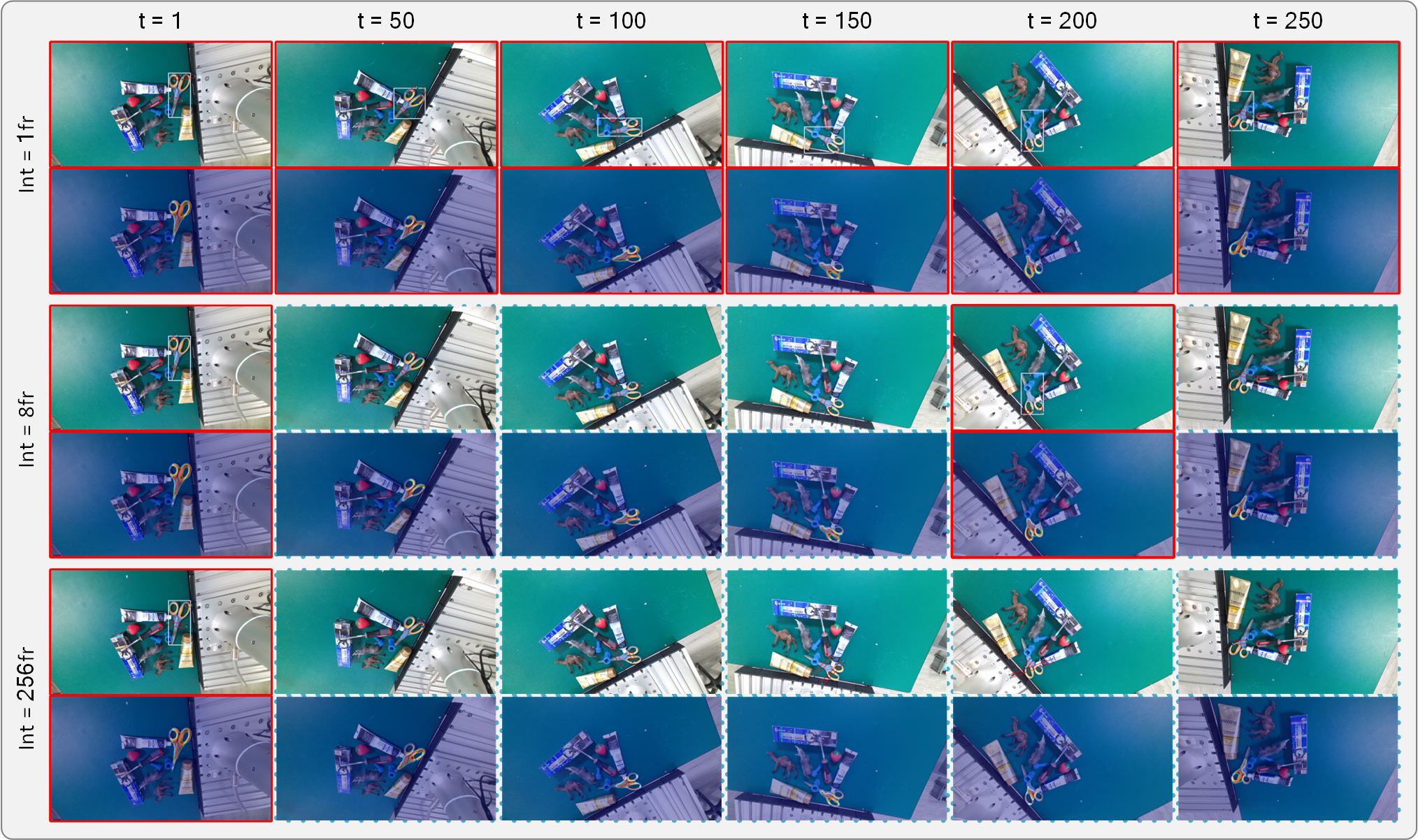}}
	\caption{Predicted grasp poses and position heatmaps at prompt intervals of $1$, $8$, and $256$ frames. Red borders indicate prompted frames, blue dashed borders denote unprompted tracking frames. Results shown for a full $256$-frame sequence, visualized at $50$-frame intervals.}
	\label{fig:grasp1}
\end{figure*}

Table~\ref{table:graspnet1} presents results for varying prompt frequencies with box prompts provided at $1$, $8$, or $256$-frame intervals. At $8$-frame intervals, SPGrasp achieved $92.5\%$, $91.7\%$, and $91.9\%$ accuracy on seen, similar, and novel object splits respectively, yielding $92.0\%$ overall accuracy. This performance surpasses RoG-SAM under comparable conditions. Performance at $1$-frame intervals is similar to $8$-frame results, while $256$-frame intervals using only first-frame prompting caused a $4.8\%$ degradation in overall accuracy. Object novelty minimally impacted SPGrasp's performance, unlike other methods showing substantial degradation on novel objects. This robustness likely stems from the foundation model's visual-language pretraining, which provides open-set visual understanding beneficial for downstream tasks, as confirmed in subsequent ablation studies. Fig. \ref{fig:grasp1} visualizes the results of inference between prompt intervals in a sequence. Frames with red borders indicate box prompt input, blue dashed borders denote unprompted tracking frames. During the first $150$ frames, all intervals show minimal differences in grasp poses and position heatmaps. However, heatmap intensity indicating prediction confidence is strongest at 1-frame intervals, moderate at $8$ frames, and weakest at $256$ frames. This aligns with expectations as box prompts serve as strong priors that influence neighbor predictions \cite{SAMAdapter}. While SPGrasp maintains persistent target tracking during long-term observation, the $256$-frame update interval may induce grasp prediction jitter, potentially leading to unstable grasp proposals. This phenomenon reduces when decreasing the prompt interval to $8$ frames, where confidence heatmap quality improves. While $1$-frame intervals ensure optimal safety, per-frame prompting incurs higher computational costs. Based on Table~\ref{table:graspnet1} and Fig.~\ref{fig:grasp1}, we selected $8$-frame intervals as a tradeoff for GraspNet-1billion experiments.

To investigate the impact of history length (\(N_\text{hist}\)) and clip length (\(N_\text{clip}\)) on performance and efficiency, we conducted an ablation study varying these hyperparameters. Table~\ref{tab:nhist_nclip_accuracy} summarizes their influence on prediction accuracy, inference speed, and GPU memory utilization during training. We report total processing time per frame for inference, which includes amortized feature extraction costs from the entire clip plus per-frame decoding and propagation time. Parenthesized values in the ``Inference Time" rows denote time consumed specifically by per-frame decoding and propagation.

Results show positive correlations between both \(N_\text{hist}\) and \(N_\text{clip}\) with model accuracy, at the cost of increased computational burden. The influence of \(N_\text{clip}\) is particularly pronounced. At \(N_\text{clip} = 1\), the model processes only single frames per scene, significantly impairing dynamic scene understanding and reducing accuracy. Increasing \(N_\text{clip}\) from $1$ to $8$ frames yields substantial accuracy improvements across all \(N_\text{hist}\) settings. For instance, with \(N_\text{hist}=8\), accuracy rises from $80.8\%$ (\(N_\text{clip}=1\)) to $92.0\%$ (\(N_\text{clip}=8\)). Further increasing \(N_\text{clip}\) from $8$ to $16$ frames provides marginal gains ($92.0\%$ to $92.1\%$ for \(N_\text{hist}=8\)) but nearly doubles training memory ($13.2$ GB to $23.1$ GB for \(N_\text{hist}=8\)). We therefore identify \(N_\text{clip}=8\) as the optimal performance-efficiency trade-off.

Inference time exhibits exclusive correlation with \(N_\text{hist}\), demonstrating an approximate $6\sim8$ ms increase per $2$-step increment in \(N_\text{hist}\) (rising from $59.4$ ms at \(N_\text{hist}=4\) to $73.1$ ms at \(N_\text{hist}=8\)), while remaining unaffected by \(N_\text{clip}\) variations. Computational analysis reveals that the initial encoding process dominates the time cost, with its expense being amortized across frames and maintaining consistency for fixed \(N_\text{hist}\) values. Only the per-frame decoding and propagation components (indicated by parenthesized values) show dependence on \(N_\text{hist}\). Remarkably, the total inference time per frame consistently remains under $100$ ms across all tested configurations, achieving a $2.4$ times speed advantage over contemporary state-of-the-art prompt-based method RoG-SAM ($176$ ms) while preserving equivalent accuracy levels.

\begin{table}[h] 
	\centering
	\caption{Model Accuracy and Inference Time for different \(N_\text{hist}\) and \(N_\text{clip}\) values.}
	\newcommand{\thicktoprule}{\toprule[1pt]}
	\label{tab:nhist_nclip_accuracy}
	\begin{tabular}{ccccc}
		\toprule
        \multirow{2}{*}{\(N_\text{hist}\)} & \multirow{2}{*}{Metric} & \multicolumn{3}{c}{\(N_\text{clip}\)} \\
        \cmidrule(lr){3-5} % Line only under N_clip columns
		 &  & {1 Frame} & {8 Frame} & {16 Frame} \\ % Replace Value A1, A2, A3 with actual N_clip
		\midrule
		\multirow{2}{*}{4} % Replace Value H1 with actual N_hist value
		& Accuracy (\%) &79.2  & 91.5 &92.1  \\
		& Inference Time (ms) &59.4(16.7) &59.4(16.7)  &59.4(16.7) \\ 
		& Memory (GB) &3.8  &12.9  &22.4  \\
		\midrule
		\multirow{2}{*}{6}
		& Accuracy (\%) &81.6  &91.7  &92.3  \\
		& Inference Time (ms) &65.3(22.9)  &65.3(22.9)  & 65.3(22.9) \\
		& Memory (GB) & 3.8   &13.0   & 22.9  \\
		\midrule
		\multirow{2}{*}{8}
		& Accuracy (\%) &80.8  & 92.0 &92.1  \\
		& Inference Time (ms) &73.1(30.4)  &73.1(30.4)  &73.1(30.4)  \\
		& Memory (GB) &3.9  &13.2  &23.1  \\
		\bottomrule
	\end{tabular}
	\end{table}

To comprehensively evaluate the performance contributions of key components in SPGrasp, we conducted an ablation study examining three critical aspects: the role of spatiotemporal memory (by removing the memory buffer), the necessity of pretraining, and the impact of SAM2 backbone model size. Results are presented in Table~\ref{tab:ablation study}. Our baseline configuration uses a SAM2-Base image encoder with pretrained weights as the reference.

The results reveal several key insights: 
\begin{itemize}
    \item \textbf{Removing spatiotemporal memory ({w/o memory buffer})} reduces inference time by $21.0\%$ (from $73.1$ ms to $57.7$ ms) but incurs a $2.6\%$ accuracy drop ($92.0\%$ to $89.4\%$).
    \item \textbf{Training from scratch ({w/o pretrained weights})} causes a substantial $15.8\%$ accuracy decrease to $76.2\%$, with unchanged inference time. This highlights the critical importance of pretrained foundation models for effective downstream task adaptation.
\end{itemize}

We further visualized tracking consistency impacts in two challenging Graspnet-1Billion scenarios (Fig.~\ref{fig:grasp2}):

\begin{figure*}[ht] 
	\centerline{\includegraphics[width=1\linewidth]{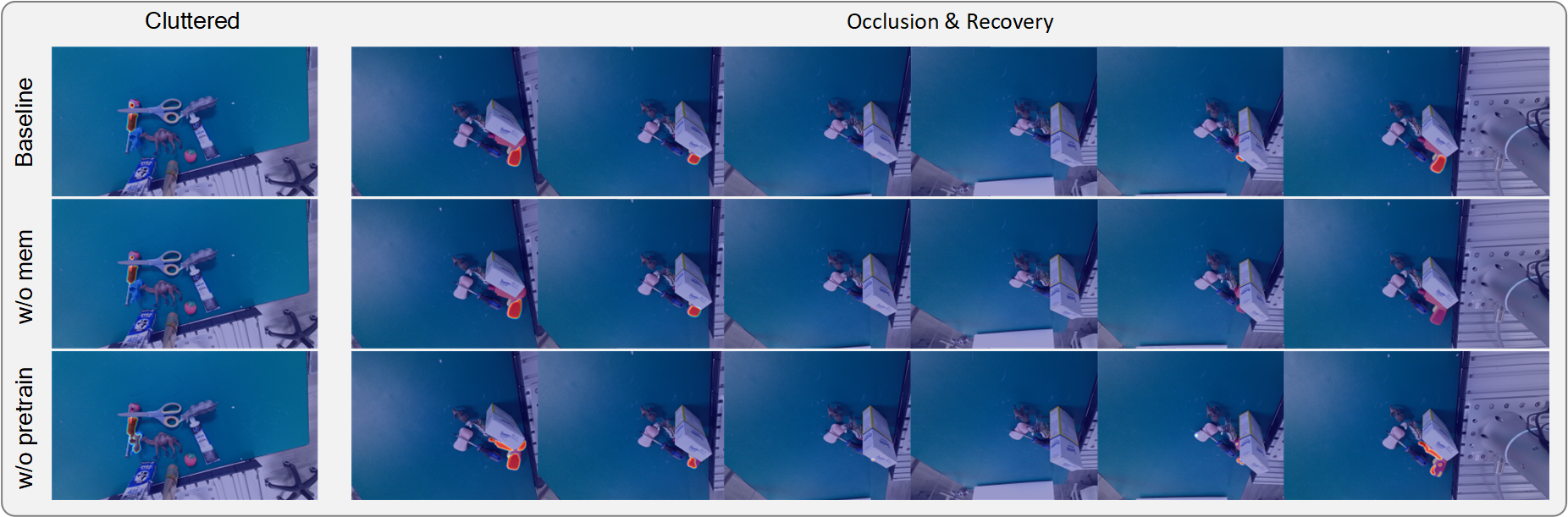}}
        \caption{Visualization of ablation study results for cluttered (single-frame) and occlusion recovery (multi-frame) scenarios in Graspnet-1Billion. For the cluttered case, prompts are provided at the target frame; for occlusion recovery, only the initial frame receives a prompt.}
	\label{fig:grasp2}
\end{figure*}

\begin{itemize}
    \item \textbf{Cluttered scenario:} Disambiguating target objects from surrounding distractors in cluttered environments presents a key challenge for prompt-driven grasping. Our experiments demonstrate this through a scenario where a screwdriver is surrounded by scissors and blue plastic models. In this setting both our baseline model and the memory-ablated configuration successfully identify the correct grasp area on the screwdriver as both leverage pretraining. Conversely the model trained from scratch fails by producing false positives on adjacent objects. This significant performance gap confirms that the rich visual understanding acquired through pretraining on large-scale video datasets like SA-V \cite{sam2} is essential for robust instance identification, a fundamental capability missing in models trained without pretraining.
    
    \item \textbf{Occlusion \& recovery:} Occlusion and recovery scenarios occur when part of a target object is blocked by another object, causing incomplete visual information. These dynamic occlusion events rigorously test model capabilities for long-term tracking and re-identification. In our experimental sequence with a red toy pistol undergoing 12 frames of occlusion before reappearing, two configurations succeeded in resuming tracking: the baseline model and the pretrain-ablated model, both containing the memory buffer. Meanwhile the memory-ablated configuration permanently lost the target. This clear performance difference confirms the memory buffer plays an essential role in maintaining target persistence and enabling recovery after occlusion.
\end{itemize}

% \vspace{5pt}
\begin{table}[ht]
	\centering
	\caption{Component ablation and backbone scaling analysis of SPGrasp on GraspNet-1Billion.} 
    \vspace{5pt}
	\newcommand{\thicktoprule}{\toprule[1pt]}
	\label{tab:ablation study}
	\begin{tabular}{l c c c} %
		\toprule % 
		Configuration & Acc. (\%) & Infer. Time (ms) & Mem. (GB) \\ 
		\midrule
		% Baseline and core component ablations
		SPGrasp (Baseline)                        & \textbf{92.0} & 73.1 & 13.2 \\ 
		\quad w/o memory buffer                    & 89.4 & \textbf{57.7} & \textbf{13.1} \\ 
		\quad w/o pretrained weights & 76.2 & 73.1 & 13.2 \\
		\midrule
		
		SPGrasp (Baseline)            & 92.0 & 73.1 & 13.2 \\
		\quad + SAM2-Small             & 87.3 & \textbf{48.6} & \textbf{6.3} \\
		\quad + SAM2-Large            & \textbf{92.5} & 148.9 & 20.8 \\
		
		\bottomrule
	\end{tabular}
\end{table}

\begin{table}[h!]
    \centering
    \vspace{3pt}
    \caption{Generalization performance of SPGrasp across different occlusion levels. The model was trained exclusively on the Unoccluded dataset.}
    \label{tab:generalization_occlusion}
    
    \begin{tabular}{lc}
        \toprule
        \textbf{Testing Scenario} & \textbf{Accuracy (\%)} \\
        \midrule
        
        Unoccluded (Seen Domain)      & \bfseries 98.7 \\ 
        \midrule 
        Partially Occluded (Unseen)   & 97.7 \\
        Heavily Occluded (Unseen)     & 94.8\\
        
        \bottomrule
    \end{tabular}
\end{table}

\begin{figure}[h!] 
	\centerline{\includegraphics[width=1\linewidth]{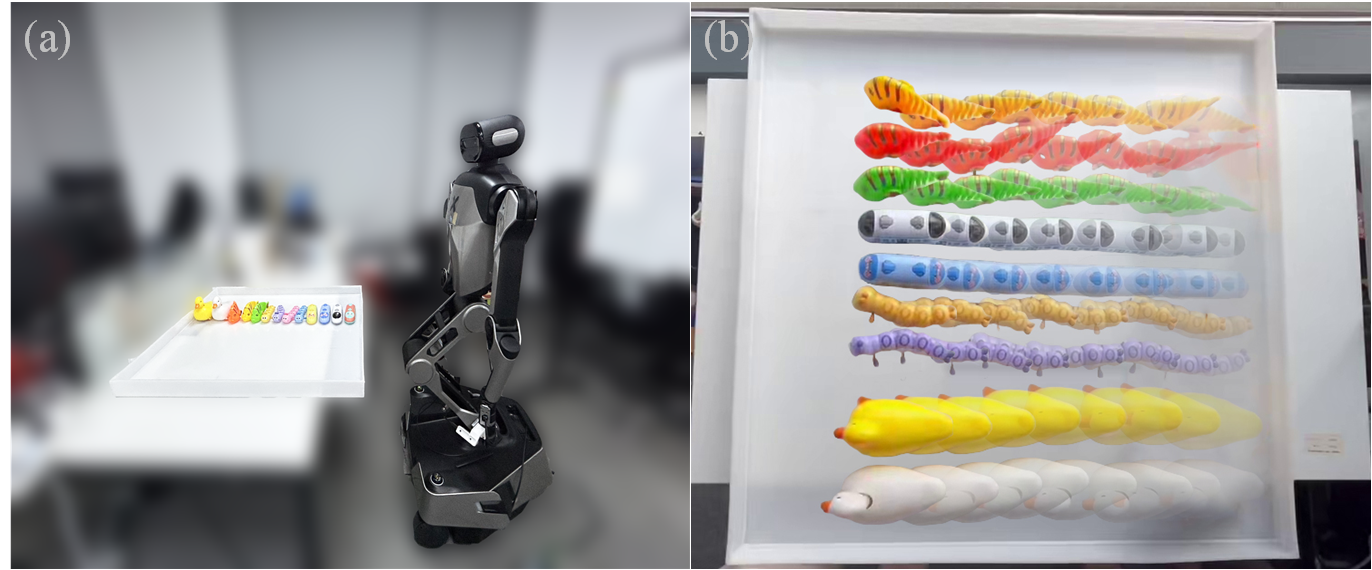}}
        \caption{The real-world experimental setup. a) The robot, camera, and operational workspace. b) The set of movable toys used as dynamic targets for grasp prediction.}
        % \vspace{-6pt}
	\label{fig:real1}
\end{figure}

Regarding backbone scaling,
\begin{itemize}
    \item \textbf{SAM2-Large substitution} achieves marginally higher accuracy ($92.5\%$) but more than doubles inference time ($148.9$ ms) and increases training memory to $20.8$ GB.
    \item \textbf{SAM2-Small} reduces inference time to $48.6$ ms and memory to $6.3$ GB, but incurs a significant $4.7\%$ accuracy penalty ($87.3\%$).
\end{itemize}

These results demonstrate trade-offs between model capacity, computational efficiency, and accuracy, justifying the baseline SAM2-Base selection.

\begin{figure*}[h!] 
	\centerline{\includegraphics[width=1\linewidth]{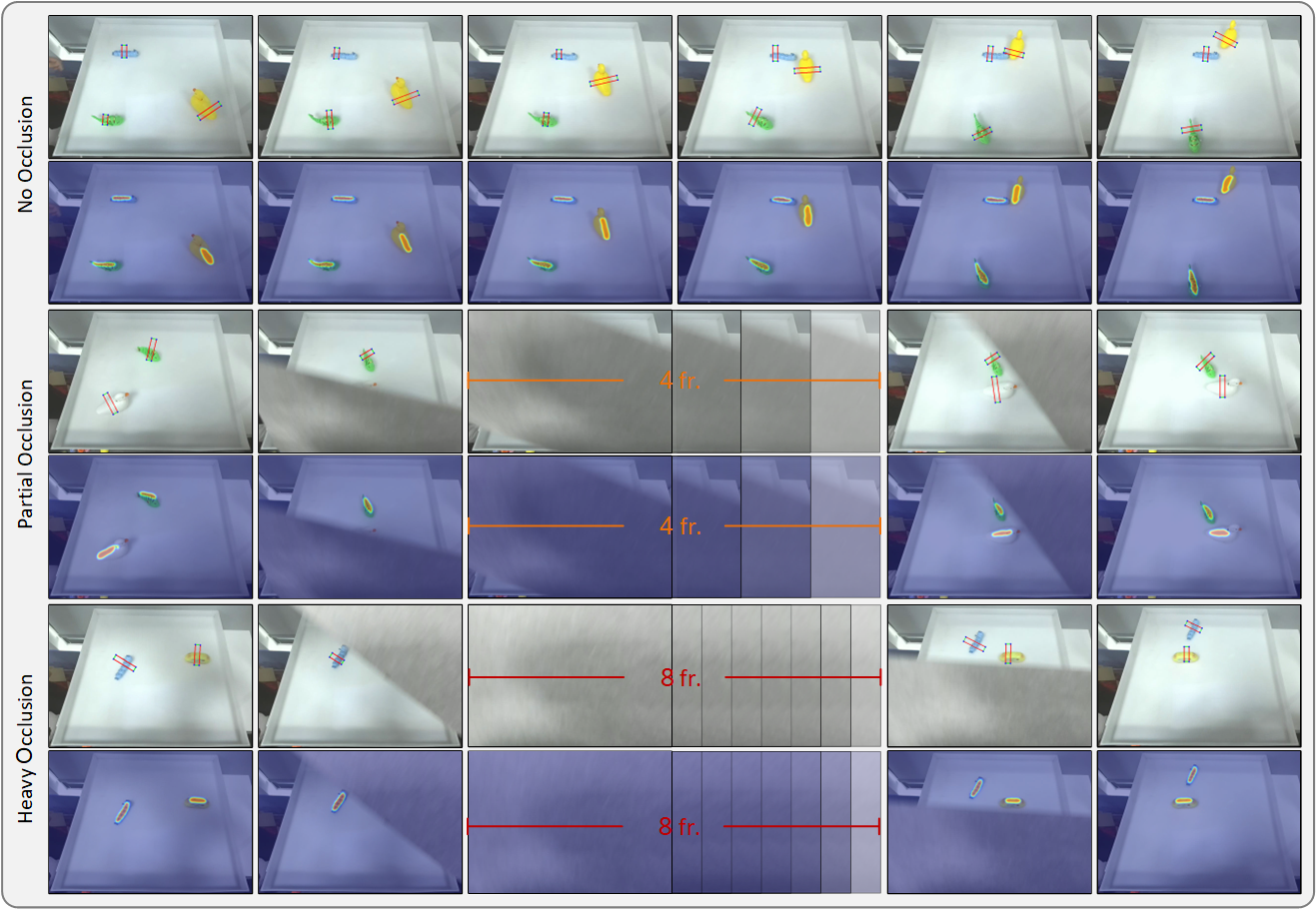}}
    \vspace{3pt}
        \caption{Qualitative results of SPGrasp's prediction under different occlusion scenarios with multiple dynamic objects. The model successfully re-establishes tracking for each target after it reappears, demonstrating the robustness endowed by its spatiotemporal memory to handle multi-instance tracking.}
	\label{fig:real2}
    \vspace{3pt}
\end{figure*}

\subsection{Real-World Experiments}
\label{sec:real_world_experiments}

To validate the practical efficacy and robustness of SPGrasp, we conducted experiments on real-world scenarios to evaluate its prediction performance in dynamic environments.

\paragraph{Experimental setup}
Our experimental setup, illustrated in Fig. \ref{fig:real1}, consists of a Galaxea R1 robot \cite{behavior}, a ZED 2i stereo camera, and two Galaxea G1 parallel gripper. The system is controlled by a workstation equipped with an NVIDIA GeForce RTX $4080 $ GPU running Ubuntu $20.04$. The prediction task involves tracking and synthesizing grasp poses for $13$ distinct movable toys as they traverse a plastic pallet within the robot's workspace.

\paragraph{Dataset and evaluation protocol}
For training, we collected a specialized dataset of $173$ unoccluded sequences featuring up to four moving objects per scene. Each sequence was downsampled to $60$ frames and manually annotated with per-frame, per-instance grasp poses and semantic masks.

To evaluate model robustness against occlusions, we designed three distinct testing scenarios: ``Unoccluded'', ``Partially Occluded'' (target obscured for $1\sim 4$ consecutive frames), and ``Heavily Occluded'' (target obscured for $5\sim 8$ consecutive frames). Occlusions were simulated by moving a baffle through the camera's field of view. For each test sequence, prompt was provided only at the first frame generated by Grounding DINO \cite{DINO}. SPGrasp was then tasked with tracking and predicting grasps for all subsequent frames without further guidance. The quantitative performance, measured by prediction accuracy across these scenarios, is detailed in Table~\ref{tab:generalization_occlusion}. Despite being trained exclusively on unoccluded data, the model achieves a accuracy of $97.7\%$ on partially occluded scenes and maintains a robust $94.8\%$ accuracy even under heavy occlusion. This minimal performance degradation highlights the effectiveness of our spatiotemporal context module, which enables the model to re-establish tracking after temporary obstructions.

We further conducted qualitative analysis of SPGrasp's performance across occlusion scenarios. Sequential visualizations in Fig. \ref{fig:real2} demonstrate stable grasp pose prediction and successful recovery tracking under three occlusion levels. Crucially, the model consistently relocalizes targets and resumes accurate prediction after full occlusion as long as the occlusion duration remains within the \(N_{\text{hist}}=8\) frame capacity of our spatiotemporal memory buffer. These results validate the memory mechanism's efficacy and confirm the model's operational readiness for robotic systems that must handle cluttered environments with intermittent occlusions.

\vspace{3pt}
\section{Conclusion}
\vspace{3pt}
This paper presents SPGrasp, a prompt-driven framework for real-time instance-level grasp synthesis in dynamic scenes. By extending SAM 2 with a spatiotemporal context module that integrates grasp memory, SPGrasp enables promptable grasp tracking with sparse user input while maintaining temporal consistency across cluttered and occluded environments. Our design eliminates continuous prompting requirements, offering high applicability in real-world settings demanding robust, low-latency interaction.
Extensive experiments on Jacquard, OCID, and GraspNet-1Billion datasets demonstrate SPGrasp achieves competitive instance-level accuracy with $2.4$ times faster inference than RoG-SAM. Real-world evaluations featuring multiple moving objects and occlusion events show high accuracy and strong generalization, even under limited fine-tuning data. These results validate SPGrasp's deployment potential for practical robotic manipulation systems.
Future work will explore extensions to full $6$-DoF grasping, integration with closed-loop control, and scaling to larger foundational vision models for further generalization. Overall, SPGrasp offers a unified and efficient solution for interactive, dynamic grasping in complex visual environments.
\vspace{8pt}
%\section{References Section}
% \newpage
\bibliography{sn-bibliography}% common bib file

\begin{thebibliography}{10}
\providecommand{\url}[1]{#1}
\csname url@samestyle\endcsname
\providecommand{\newblock}{\relax}
\providecommand{\bibinfo}[2]{#2}
\providecommand{\BIBentrySTDinterwordspacing}{\spaceskip=0pt\relax}
\providecommand{\BIBentryALTinterwordstretchfactor}{4}
\providecommand{\BIBentryALTinterwordspacing}{\spaceskip=\fontdimen2\font plus
\BIBentryALTinterwordstretchfactor\fontdimen3\font minus
  \fontdimen4\font\relax}
\providecommand{\BIBforeignlanguage}[2]{{%
\expandafter\ifx\csname l@#1\endcsname\relax
\typeout{** WARNING: IEEEtran.bst: No hyphenation pattern has been}%
\typeout{** loaded for the language `#1'. Using the pattern for}%
\typeout{** the default language instead.}%
\else
\language=\csname l@#1\endcsname
\fi
#2}}
\providecommand{\BIBdecl}{\relax}
\BIBdecl

\bibitem{GRConv}
S.~Kumra, S.~Joshi, and F.~Sahin, ``Antipodal robotic grasping using generative
  residual convolutional neural network,'' in \emph{Proc. IEEE/RSJ Int. Conf.
  Intell. Robots Syst.}\hskip 1em plus 0.5em minus 0.4em\relax IEEE, 2020, pp.
  9626--9633.

\bibitem{feng2025multi}
Z.~Feng, R.~Xue, L.~Yuan, Y.~Yu, N.~Ding, M.~Liu, B.~Gao, J.~Sun, X.~Zheng, and
  G.~Wang, ``Multi-agent embodied {AI: A}dvances and future directions,''
  \emph{arXiv:2505.05108}, 2025.

\bibitem{sam2}
N.~Ravi, V.~Gabeur, Y.-T. Hu, R.~Hu, C.~Ryali, T.~Ma, H.~Khedr, R.~R{\"a}dle,
  C.~Rolland, L.~Gustafson \emph{et~al.}, ``{SAM} 2: {S}egment anything in
  images and videos,'' \emph{arXiv:2408.00714}, 2024.

\bibitem{Tip}
X.~Feng, Y.~Li, D.~Chen, C.~Qiao, J.~Yuan, L.~Yuan, and G.~Hua, ``Pluralistic
  salient object detection,'' \emph{IEEE Trans. Image Process.}, 2025.

\bibitem{Nips}
Y.~Zhai, K.~Lin, Z.~Yang, L.~Li, J.~Wang, C.-C. Lin, D.~Doermann, J.~Yuan, and
  L.~Wang, ``Motion consistency model: Accelerating video diffusion with
  disentangled motion-appearance distillation,'' \emph{Adv. Neural Inf.
  Process. Syst.}, vol.~37, pp. 111\,000--111\,021, 2024.

\bibitem{GraspAnything2}
A.~Sahbani, S.~El-Khoury, and P.~Bidaud, ``An overview of {3D} object grasp
  synthesis algorithms,'' \emph{Robot. Auton. Syst.}, vol.~60, no.~3, pp.
  326--336, 2012.

\bibitem{universal}
Y.~Xie, M.~Li, S.~Li, X.~Li, G.~Chen, F.~Ma, F.~R. Yu, and W.~Ding, ``Universal
  visuo-tactile video understanding for embodied interaction,''
  \emph{arXiv:2505.22566}, 2025.

\bibitem{manigaussian++}
T.~Yu, G.~Lu, Z.~Yang, H.~Deng, S.~S. Chen, J.~Lu, W.~Ding, G.~Hu, Y.~Tang, and
  Z.~Wang, ``{ManiGaussian++: G}eneral robotic bimanual manipulation with
  hierarchical {G}aussian world model,'' \emph{arXiv:2506.19842}, 2025.

\bibitem{6dtrack}
B.~Burgess-Limerick, C.~Lehnert, J.~Leitner, and P.~Corke, ``Dgbench: An
  open-source, reproducible benchmark for dynamic grasping,'' in \emph{Proc.
  IEEE/RSJ Int. Conf. Intell. Robots Syst.}\hskip 1em plus 0.5em minus
  0.4em\relax IEEE, 2022, pp. 3218--3224.

\bibitem{motion}
N.~Chen, X.-M. Wu, G.~Xu, J.-J. Jiang, Z.~Chen, and W.-S. Zheng,
  ``{MotionGrasp: L}ong-term grasp motion tracking for dynamic grasping,''
  \emph{IEEE Robot. Autom. Lett.}, vol.~10, no.~1, pp. 796--803, 2025.

\bibitem{M2}
J.~Liu, R.~Zhang, H.-S. Fang, M.~Gou, H.~Fang, C.~Wang, S.~Xu, H.~Yan, and
  C.~Lu, ``Target-referenced reactive grasping for dynamic objects,'' in
  \emph{Proc. IEEE/CVF Conf. Comput. Vis. Pattern Recognit.}, 2023, pp.
  8824--8833.

\bibitem{Anygrasp}
H.-S. Fang, C.~Wang, H.~Fang, M.~Gou, J.~Liu, H.~Yan, W.~Liu, Y.~Xie, and
  C.~Lu, ``{Anygrasp: R}obust and efficient grasp perception in spatial and
  temporal domains,'' \emph{IEEE Trans. Robot.}, vol.~39, no.~5, pp.
  3929--3945, 2023.

\bibitem{SAM}
A.~Kirillov, E.~Mintun, N.~Ravi, H.~Mao, C.~Rolland, L.~Gustafson, T.~Xiao,
  S.~Whitehead, A.~C. Berg, W.-Y. Lo \emph{et~al.}, ``Segment anything,'' in
  \emph{Proc. IEEE/CVF Int. Conf. Comput. Vis.}, 2023, pp. 4015--4026.

\bibitem{rog-sam}
Y.~Mei, J.~Sun, Z.~Peng, F.~Deng, G.~Wang, and J.~Chen, ``{RoG-SAM: A}
  language-driven framework for instance-level robotic grasping detection,''
  \emph{IEEE Trans. Multimedia}, vol.~27, pp. 3057--3068, 2025.

\bibitem{GGCNN}
D.~\vspace{0mm}Morrison, P.~Corke, and J.~Leitner, ``Closing the loop for
  robotic grasping: {A} real-time, generative grasp synthesis approach,''
  \emph{arXiv:1804.05172}, 2018.

\bibitem{Graspnet}
H.-S. Fang, C.~Wang, M.~Gou, and C.~Lu, ``{GraspNet-1Billion: A} large-scale
  benchmark for general object grasping,'' in \emph{Proc. IEEE/CVF Conf.
  Comput. Vis. Pattern Recognit.}, 2020, pp. 11\,444--11\,453.

\bibitem{M2.1}
I.~Akinola, J.~Xu, S.~Song, and P.~K. Allen, ``Dynamic grasping with
  reachability and motion awareness,'' in \emph{Proc. IEEE/RSJ Int. Conf.
  Intell. Robots Syst.}\hskip 1em plus 0.5em minus 0.4em\relax IEEE, 2021, pp.
  9422--9429.

\bibitem{VFAS24}
P.~Rosenberger, A.~Cosgun, R.~Newbury, J.~Kwan, V.~Ortenzi, P.~Corke, and
  M.~Grafinger, ``Object-independent human-to-robot handovers using real time
  robotic vision,'' \emph{IEEE Robot. Autom. Lett.}, vol.~6, no.~1, pp. 17--23,
  2021.

\bibitem{yolov3}
J.~Redmon and A.~Farhadi, ``{Yolov3: An} incremental improvement,''
  \emph{arXiv:1804.02767}, 2018.

\bibitem{liu2023edgeyolo}
S.~Liu, J.~Zha, J.~Sun, Z.~Li, and G.~Wang, ``{EdgeYOLO: A}n edge-real-time
  object detector,'' in \emph{Chinese Control Conf.}\hskip 1em plus 0.5em minus
  0.4em\relax IEEE, 2023, pp. 7507--7512.

\bibitem{CLIP}
A.~Radford, J.~W. Kim, C.~Hallacy, A.~Ramesh, G.~Goh, S.~Agarwal, G.~Sastry,
  A.~Askell, P.~Mishkin, J.~Clark \emph{et~al.}, ``Learning transferable visual
  models from natural language supervision,'' in \emph{Proc. Int. Conf. Mach.
  Learn.}\hskip 1em plus 0.5em minus 0.4em\relax PMLR, 2021, pp. 8748--8763.

\bibitem{LanguageGrasp}
G.~Tziafas, Y.~Xu, A.~Goel, M.~Kasaei, Z.~Li, and H.~Kasaei, ``Language-guided
  robot grasping: {C}lip-based referring grasp synthesis in clutter,''
  \emph{arXiv:2311.05779}, 2023.

\bibitem{prompt2}
L.~Barcellona, A.~Bacchin, M.~Terreran, E.~Menegatti, and S.~Ghidoni, ``Show
  and grasp: {F}ew-shot semantic segmentation for robot grasping through
  zero-shot foundation models,'' \emph{arXiv:2404.12717}, 2024.

\bibitem{DINO}
H.~Zhang, F.~Li, S.~Liu, L.~Zhang, H.~Su, J.~Zhu, L.~M. Ni, and H.-Y. Shum,
  ``Dino: Detr with improved denoising anchor boxes for end-to-end object
  detection,'' \emph{arXiv:2203.03605}, 2022.

\bibitem{GraspRPN}
H.~Zhang, X.~Zhou, X.~Lan, J.~Li, Z.~Tian, and N.~Zheng, ``A real-time robotic
  grasping approach with oriented anchor box,'' \emph{IEEE Trans. Syst. Man
  Cybern. Syst.}, vol.~51, no.~5, pp. 3014--3025, 2019.

\bibitem{OCID}
S.~Ainetter and F.~Fraundorfer, ``{End-to-end trainable deep neural network for
  robotic grasp detection and semantic segmentation from RGB},'' in \emph{Proc.
  IEEE Int. Conf. Robot. Autom.}\hskip 1em plus 0.5em minus 0.4em\relax IEEE,
  2021, pp. 13\,452--13\,458.

\bibitem{Jacquard}
A.~Depierre, E.~Dellandr{\'e}a, and L.~Chen, ``Jacquard: {A} large scale
  dataset for robotic grasp detection,'' in \emph{Proc. IEEE/RSJ Int. Conf.
  Intell. Robots Syst.}, 2018, pp. 3511--3516.

\bibitem{Ins}
Y.~Xu, M.~Kasaei, H.~Kasaei, and Z.~Li, ``Instance-wise grasp synthesis for
  robotic grasping,'' in \emph{Proc. IEEE Int. Conf. Robot. Autom.}\hskip 1em
  plus 0.5em minus 0.4em\relax IEEE, 2023, pp. 1744--1750.

\bibitem{multigrasp}
F.-J. Chu, R.~Xu, and P.~A. Vela, ``Real-world multiobject, multigrasp
  detection,'' \emph{IEEE Robot. Autom. Lett.}, vol.~3, no.~4, pp. 3355--3362,
  2018.

\bibitem{Morrison}
D.~Morrison, P.~Corke, and J.~Leitner, ``Learning robust, real-time, reactive
  robotic grasping,'' \emph{Int. J. Robot. Res.}, vol.~39, no. 2-3, pp.
  183--201, 2020.

\bibitem{MSGNet}
S.~Duan, G.~Tian, Z.~Wang, S.~Liu, and C.~Feng, ``A semantic robotic grasping
  framework based on multi-task learning in stacking scenes,'' \emph{Eng. Appl.
  Artif. Intell.}, vol. 121, p. 106059, 2023.

\bibitem{GRConv2}
S.~Kumra, S.~Joshi, and F.~Sahin, ``{GR-ConvNet v2: A} real-time multi-grasp
  detection network for robotic grasping,'' \emph{Sensors}, vol.~22, no.~16, p.
  6208, 2022.

\bibitem{SAMAdapter}
J.~Wu, W.~Ji, Y.~Liu, H.~Fu, M.~Xu, Y.~Xu, and Y.~Jin, ``Medical {SAM} adapter:
  {A}dapting segment anything model for medical image segmentation,''
  \emph{arXiv:2304.12620}, 2023.

\bibitem{behavior}
Y.~Jiang, R.~Zhang, J.~Wong, C.~Wang, Y.~Ze, H.~Yin, C.~Gokmen, S.~Song, J.~Wu,
  and L.~Fei-Fei, ``Behavior robot suite: {S}treamlining real-world whole-body
  manipulation for everyday household activities,'' \emph{arXiv:2503.05652},
  2025.

\end{thebibliography}
\newpage


\vfill

\end{document}